\pdfoutput=1

\documentclass[11pt]{article}

\usepackage{emnlp2021}

\usepackage{times}
\usepackage{latexsym}

\usepackage[T1]{fontenc}

\usepackage[utf8]{inputenc}

\usepackage{microtype}
\usepackage{graphicx}
\usepackage{amsmath}
\usepackage{subcaption}

\title{Detecting Polarized Topics Using Partisanship-aware Contextualized Topic Embeddings}

\author{
  Zihao He$^{1,4}$, Negar Mokhberian$^{1,4}$, Ant\'onio C\^amara$^{2}$, Andr\'es Abeliuk$^{3}$, Kristina Lerman$^{4}$ \\
  $^1$Department of Computer Science, University of Southern California \\
  $^2$Department of Computer Science, Columbia University \\
  $^3$Department of Computer Science, University of Chile\\
  $^4$Information Sciences Institute, University of Southern California\\
  \texttt{ \{zihaoh,nmokhber\}@usc.edu}   \quad  \texttt{a.camara@columbia.edu} \\
  \texttt{aabeliuk@dcc.uchile.cl} \quad  \texttt{lerman@isi.edu}\\
  }

\begin{document}
\maketitle
\begin{abstract}
Growing polarization of the news media has been blamed for fanning disagreement, controversy and even violence. Early identification of polarized topics is thus an urgent matter that can help mitigate conflict. However, accurate measurement of topic-wise polarization is still an open research challenge. To address this gap, we propose \textbf{P}artisanship-\textbf{a}ware \textbf{C}ontextualized \textbf{T}opic \textbf{E}mbeddings (\textbf{PaCTE}), a method to automatically detect polarized topics from partisan news sources. Specifically, utilizing a language model that has been finetuned on recognizing partisanship of the news articles, we represent the ideology of a news corpus on a topic by \emph{corpus-contextualized topic embedding} and measure the polarization using cosine distance. We apply our method to a dataset of news articles about the COVID-19 pandemic. Extensive experiments on different news sources and topics demonstrate the efficacy of our method to capture topical polarization, as indicated by its effectiveness of retrieving the most polarized topics.\footnote{Code and data are publicly available at \url{https://github.com/ZagHe568/pacte-polarized-topics-detection}.}
\end{abstract}

\section{Introduction}
The media environment has grown increasingly polarized in recent years, creating social, cultural and political divisions~\cite{prior2013media, fiorina2008political}. Although a diversity of opinions is healthy, and even necessary for democratic discourse, unchecked polarization can paralyze society by suppressing consensus required for effective governance \cite{tworzecki2019poland}. In more extreme cases, polarization leads to disagreement, conflict and even violence. The COVID-19 pandemic has exposed many of our vulnerabilities to the pernicious effects of polarization. Public opinions about COVID-19 \cite{jiang2020political}, as well as messaging by political elites \cite{green2020elusive, bhanot2020partisan}, are sharply divided along partisan lines.  According to a Pew Report \cite{pew2020}, partisanship significantly explains attitudes about the costs and benefits of various mitigation strategies, including non-pharmaceutical interventions and lockdowns, and even explains regional differences in the pandemic's toll in the US \cite{gollwitzer2020partisan}. 

In mass media a variety of topics is discussed every day, and polarization can form on different topics. Therefore, identifying nascent disagreements and growing controversies of different topics in news media and public discourse would help journalists craft more balanced news coverage \cite{lorenz2020behavioural, chen2020analyzing}.
Different from previous works that study polarization from a more coarse-grained perspective, \citet{demszky2019analyzing} were the first to study polarized topics using tweets about 21 mass shootings to show that some topics were more polarized than others. However, their approach to represent semantic information with word frequencies is less expressive than modern methods allow.

To better capture the topical polarization among partisan (liberal vs. conservative) media sources,
we propose \textbf{P}artisanship-\textbf{a}ware \textbf{C}ontextualized \textbf{T}opic \textbf{E}mbeddings (\textbf{PaCTE}). Specifically, given a text corpus containing news articles from both sides, we first extract a set of topics utilizing LDA topic modeling \cite{blei2003latent}. Next, we finetune a pretrained language model \cite{devlin2018bert} to recognize the partisanship of the news articles so as to render it \emph{partisanship-aware}. Then for each article, we represent its ideology on a topic by a vector, called document-contextualized\footnote{We use ``article'' and ``document'' interchangeably.} (DC) topic embedding, by aggregating language model representations of the topic keywords contextualized by the article. Such a representation sheds light primarily on the tokens that appear in the topic keywords and thus concentrates on the topic-oriented local semantics in the context of the article, instead of the global semantics from the article that might contain irrelevant and noisy information. We further represent the ideology of the news corpus on the topic, what we call corpus-contextualized (CC) topic embedding, by aggregating the DC topic embeddings. As a result, the ideology of the news corpus on a topic is represented by a single vector. Finally, we measure the polarization between two news sources on the topic using the cosine distance between such vectors. 

For evaluation, we create ground truth by annotating the polarization of pairs of partisan news sources on a variety of topics. We evaluate the topic polarization scores produced by PaCTE against the ground truth on the task of polarized topics retrieval. Experiments on nine pairs of partisan news sources demonstrate that compared to baselines, PaCTE is more effective in capturing topic polarization and retrieving polarized topics. We argue that public media watchdogs and social media platforms can utilize such a simple-yet-effective tool to flag discussions that have grown divisive so that action could be taken to reduce partisan divisions and improve civil discourse. 


\section{Related Work}
The partisan polarization in the US media is a widely studied topic \cite{hollander2008tuning, stroud2011niche}. During the onset of the COVID-19 pandemic, the polarization among the political elites and the news media causes a lot of confusion. For example, \citet{hart2020covid19} show that COVID-19 media coverage is politicized and polarized. Other works have been studying the polarization in media from different perspectives. Focusing on the differences in the languages of liberals and conservatives, \citet{khudabukhsh2020we} analyze political polarization on YouTube using machine translation tools. To analyze how the news outlets frame the events differently, \citet{fan2019plain} have collected and labeled 100 triplets of news articles each discussing the same event from three news sources bearing different political ideologies.

In addition to qualitatively analyzing polarization, different approaches to quantifying polarization have also been proposed. 
\citet{gentzkow2019measuring} propose two different ways, namely the leave-out estimator and the multinomial regression, to measure the trends of partisanship in congressional speech. \citet{green2020elusive} define the polarization as one’s ability to identify the partisanship of a tweet’s author based on the contents of tweets and investigate the polarization regarding COVID-19 among political elites on Twitter. \citet{demszky2019analyzing} first measure topic-wise polarization using the leave-out estimator proposed by \cite{gentzkow2019measuring}; however, they use a token frequency vector to represent an article, which is less expressive and fails to make use of the rich semantics in the context and the pre-knowledge in
pretrained language models \cite{devlin2018bert, liu2019roberta} or pretrained word embeddings \cite{mikolov2013distributed, pennington2014glove}; furthermore, they represent the topic using the token frequency vector of the entire document, thus incurring noisy information that might smooth over the target semantics in the locality of topic keywords. In contrast, our method represents the topic embedding in the context of a document, thus generating topic representations with more attention to the target topic keywords as well as making use of the contextualized semantics from the document, as captured by the contextualized embeddings.

Some works have proposed contextualized embeddings to enhance the quality of neural topic models \cite{bianchi2020pre, chaudhary2020topicbert}. However, the scope of this work is to generate better contextualize topic embeddings for articles to capture topic polarization, with a given topic model; the exploration of other topic modeling techniques is beyond the scope of this work. 


\section{Methodology}
\label{sec:method}
\begin{figure*}[ht]
    \centering
    \includegraphics[width=0.9\textwidth]{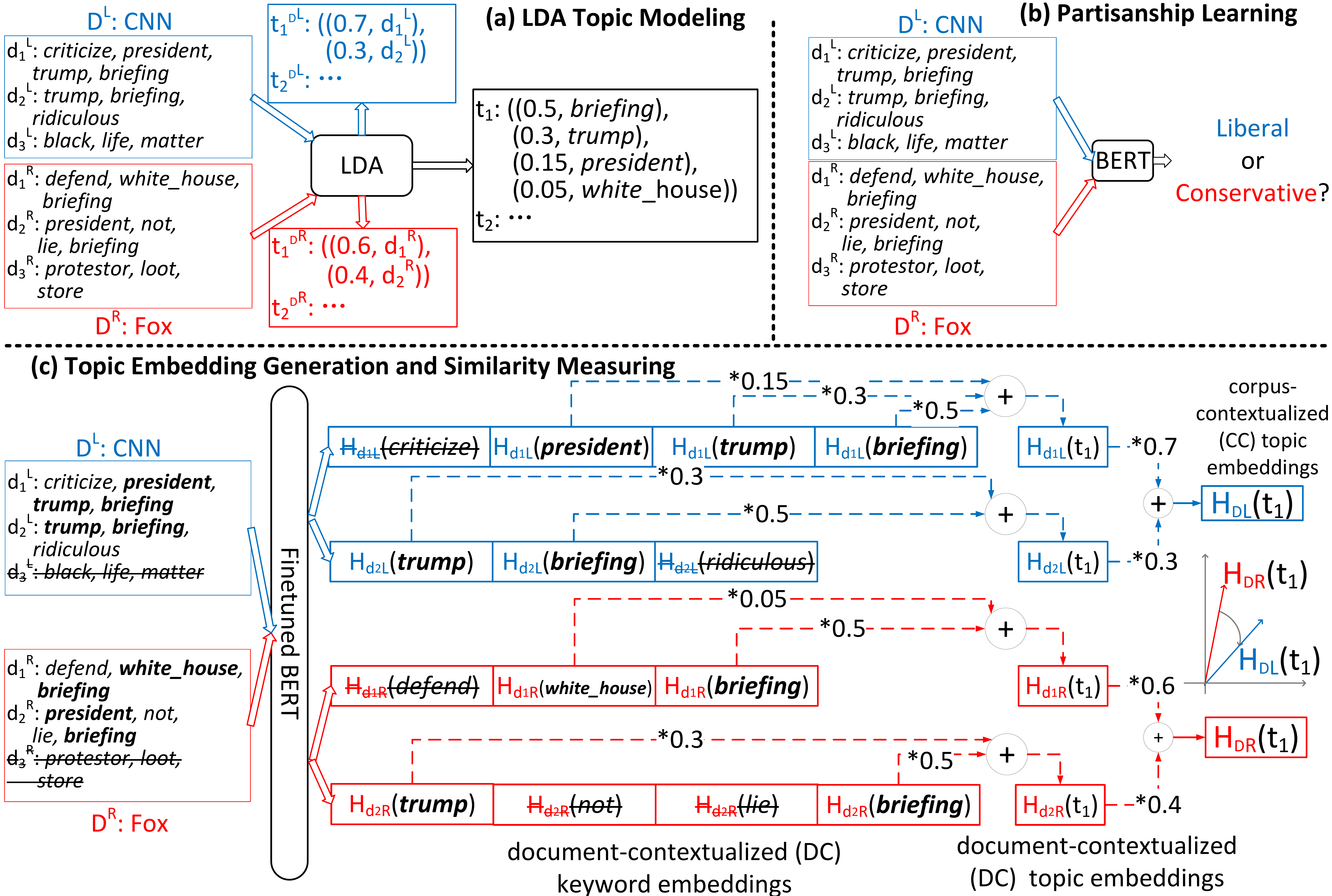}
    \caption{Overview of our PaCTE framework to detect polarized topics in media, illustrated by a toy example on CNN vs. Fox, both consisting 3 documents.
    \textbf{(a) \textit{LDA topic modeling.}} 
    We train an LDA model on the combined corpus and extract 2 topics. 
    Top-$4$ keywords on topic $t_1$ are ``\emph{briefing}'', ``\emph{trump}'', ``\emph{president}'' and ``\emph{white\_house}''. Top-$2$ most relevant documents on topic $t_1$ are $d_1^L$ and $d_2^L$ for CNN and $d_1^R$ and $d_2^R$ for Fox. $d_3^L$ and $d_3^R$ are not among the most relevant documents of this topic and are excluded in the embedding generation step. Note that we set $K=2$ (No. of topics), $m=4$ (No. of keywords), and $n=2$ (No. of documents), just for clear demonstration.
    \textbf{(b) \textit{Partisanship learning.}}
    We finetune a pretrained language model to classify the partisanship (liberal vs. conservative) of input documents.
    \textbf{(c) \textit{Topic embedding generation and similarity measuring.}} 
    We provide a step by step illustration of DC keyword embedding $\rightarrow$ DC topic embedding $\rightarrow$ CC topic embedding on topic $t_1$.
    In the two input corpora, the tokens that are among the top-$4$ keywords of topic $t_1$ are highlighted in bold.
    Take document $d_1^L$ from CNN as an example. The weighted average of the DC keyword embeddings ($H_{d_1^L}(\text{\emph{president}})$, $H_{d_1^L}(\text{\emph{trump}})$, and $H_{d_1^L}(\text{\emph{briefing}})$) is defined as the DC topic embedding $H_{d_1^L}(t_1)$ with keyword coefficients given by Eq. \ref{eq:topic2keyword}; note that $H_{d_1^L}(\text{\emph{criticize}})$ is excluded because ``\emph{criticize}'' is not among the top-$4$ keywords of topic $t_1$.
    Similarly we can obtain the DC topic embeddings for $d_2^L$, $d_1^R$ and $d_2^R$.
    The DC topic embeddings are further aggregated into CC topic embeddings $H_{D^L}(t_1)$ and $H_{D^R}(t_1)$ (document coefficients are from Eq. \ref{eq:topic2doc}) and the cosine distance between them is used as a measure of polarization of the two corpora on topic $t_1$.
    }
    \label{fig:framework_overview}
\end{figure*}

The proposed PaCTE framework consists of four components: 1) LDA Topic Modeling, 2) Partisanship Learning, 3) Partisanship-aware Contextualized Topic Embedding Generation, and 4) Measuring Polarization and Ranking Topics. The overall framework is illustrated in Figure~\ref{fig:framework_overview}. In this section we elaborate on each component in detail.  

\subsection{Problem Definition}
The input is a liberal news corpus $D^L=\{d^L_i\}_{i=1}^{|D^L|}$ and a conservative news corpus $D^R=\{d^R_i\}_{i=1}^{|D^R|}$ (L denotes "Left" and R denotes "Right"), where $d^L_i$ is an article from $D^L$ and $d^R_i$ is an article from $D^R$. A news article is represented as a sequence of tokens: $d_k=(w_i^k)_{i=1}^{|d_k|}$.
Given a topic model trained on the combined corpus $D^C=D^L \cup D^R$ with a set of modeled topics $T=\{t_i\}_{i=1}^K$ where $t_i$ represents a topic, we aim to learn a model $f$ that is able to detect the topic polarization between $D^L$ and $D^R$ on topics in $T$ and output a ranking of topics based on polarization, such that
\begin{equation}
\begin{aligned}
    & f(D^L, D^R, T)=(t_k)_{k=1}^K,\\
    & i>j \Leftrightarrow \beta(t_i, D^L, D^R)<\beta(t_j, D^L, D^R),
\end{aligned}
\end{equation}
where $\beta(t,D^L,D^R)$ represents the polarization score of topic $t$ between $D^L$ and $D^R$.

\subsection{LDA Topic Modeling}
\label{sec:lda}
We train an LDA topic model using the the combined corpus $D^C=D^L \cup D^R$ and extract $K$ topics $T=\{t_i\}_{i=1}^K$, where $t_i$ is a topic. The modeled topics $T$ apply to both $D^L$ and $D^R$. An example is given in Figure \ref{fig:framework_overview}(a).

\textbf{Representing a topic by keywords.} A topic $t_i$ is represented as a distribution of keywords from the global vocabulary of $D^C$ and we only keep the top-$m$ keywords:
\begin{equation}
\label{eq:topic2keyword}
    t_i = ((p_{ij}, w_j))_{j=1}^{m}, p_{ij}>p_{ik} \Leftrightarrow j<k,
\end{equation}
where $p_{ij}$ is the probability of observing keyword $w_j$ given topic $t_i$.

\textbf{Representing a topic by documents.} 
A document $d \in D^C$ is represented as a distribution over the $K$ topics. 
Accordingly, we renormalize the probabilities and represent each topic $t_i$ as an (inverse) distribution of documents in $D^C$ and only keep the top-$n$ most relevant documents, such that
\begin{equation}
\label{eq:topic2doc}
\scalebox{0.95}{$t_i^{D^C} = ( (q_{ij}^{D^C}, d_j) )_{j=1}^n, q_{ij}^{D^C}>q_{ik}^{D^C} \Leftrightarrow j<k,$}
\end{equation}
where $q_{ij}^D$ is the probability of observing document $d_j \in D^C$ given topic $t_i$. 
Because our goal is to study the polarization between $D^L$ and $D^R$, instead of using the global documents in $D^C$, we represent a topic by the top-$n$ documents in $D^L$ and $D^R$ separately and thus obtain $t_i^{D^L}$ and $t_i^{D^R}$ accordingly.

\subsection{Learning Partisanship}
\label{sec:learn-political-learning}
As we will see in Sections \ref{sec:emb-gene} and \ref{sec:measuring_pola}, the contextualized topic embeddings are generated from a pretrained language model \cite{devlin2018bert} and cosine distance between the topic embeddings from two corpora are used as a measure of topic polarization. The idea is inspired by static word embedding models like GloVe \cite{pennington2014glove}, where the authors measure the similarity between words by the cosine similarity between the word embeddings. 

However, to apply this measure of similarity, the model should be fitted on the target corpus. To fit the pretrained language model on the news corpora, we can use one of the two training tasks: masked language modeling or partisanship recognition. We decide on the second task because 1) it is more time efficient; 2) it informs the language model of the partisan divisions between different news sources, enhancing the language model's ability to encode the polarization arising from partisan differences in its output. This idea is similar to \cite{webson2020undocumented} where the authors call the embedding space of the language model after finetuning as ``connotation space''.
As a result, given a document $d \in D^C$, the model is optimized to classify whether it is from $D^L$ or $D^R$ by a binary cross-entropy loss, where the [CLS] embedding is used to represent the document, as shown in Figure \ref{fig:framework_overview}(b).

\subsection{Partisanship-aware Contextualized Topic Embedding Generation}
\label{sec:emb-gene}
Denote the ideology embedding of $A$ on $B$ as $H_A(B)$, where $A$ represents a news corpus or a document and $B$ represents a topic or a topic keyword.
We then represent the ideology of a corpus $D$ on a topic $t$ as corpus-contextualized (CC) topic embedding $H_D(t)$, the ideology of a document $d$ on a topic $t$ as document-contextualized (DC) topic embedding $H_d(t)$, and the ideology of a document $d$ on a topic keyword $w$ as DC keyword embedding $H_d(w)$. We will elaborate on how the CC topic embedding is obtained from a top-down perspective.

According to Equation~\ref{eq:topic2doc}, in order to compute the CC topic embedding $H_{D}(t_i)$, we can rewrite it as 
\begin{equation}
H_{D}(t_i) = \sum_{j=1}^{n} q_{ij}^D H_{d_j}(t_i).
\end{equation}
Hence, we decompose a CC topic embedding into DC topic embeddings from the top-$n$ most relevant documents.

To obtain the DC topic embedding, \citet{demszky2019analyzing} use word frequency vectors; \citet{grootendorst2020bertopic} takes the [CLS] embedding of a pretrained language model that gives a holistic document embedding without encoding the context of a topic.
However, while word frequency vectors encode statistical features of words in the document, they neglect their context. In addition, a document is likely to be associated with multiple topics according to the LDA topic model, and therefore using the holistic document embedding as the topic embedding regardless of the specific topic results in identical embeddings for different topics on the same document; moreover, even if a document is only associated with one topic, it might contain information not relevant to that topic and thus the holistic document embedding will encode noisy information. Therefore, we argue that the DC topic embedding should be both contextualized and topic-specific. In this regard, according to Equation \ref{eq:topic2keyword}, we rewrite the DC topic embedding as the weighted sum of DC keyword embeddings where only top-$m$ topic keywords are used instead of all the words in the document, as
\begin{equation}
\label{eq:topic2token}
    H_{d_j}(t_i) = \sum_{k=1}^{m} p_{ik}H_{d_j}(w_k).
\end{equation}

Finally, in terms of the DC keyword embedding $H_{d_j}(w_k)$, as can be told from its name, it is precisely what a pretrained language model \cite{devlin2018bert} is designed for. Therefore, we take the corresponding final-layer token embedding of $w_k$ when the input to the language model is $d_j$. Due to the self-attention mechanism \cite{vaswani2017attention} in the pretrained language model, $H_{d_j}(t_i)$ encodes the global context of the document, but since it only takes the sum of topic keyword embeddings, the encoded information is more oriented towards this specific topic $t_i$, which elegantly resonates with its name ``document-contextualized topic embedding''.
The step-by-step illustration of the generation of $H_d(w)$, $H_d(t)$ and $H_D(t)$ is shown in Figure \ref{fig:framework_overview} (c).

Because the language model used to generate the embeddings is finetuned to encode partisanship, the generated $H_{D}(t_i)$ also contains this information and is more precisely called partisanship-aware corpus-contextualized topic embedding. For brevity we call it corpus-contextualized (CC) topic embedding.

\subsection{Measuring Polarization and Ranking Topics}
\label{sec:measuring_pola}
After obtaining the CC topic embeddings $H_{D^L}(t_i)$ and $H_{D^R}(t_i)$ of the two corpora $D^L$ and $D^R$ on topic $t_i$, using two different sets of top-$n$ most relevant documents from $D^L$ and $D^R$ respectively, we measure the ideology similarity (and then polarization) based on the cosine similarity between them, such that

\begin{equation}
\begin{aligned}
& c  = \text{cos\_sim}(H_{D^L}(t_i), H_{D^R}(t_i)), \\
& \beta(D^L, D^R, t_i) = 0.5*(1-c) \in [0,1].
\end{aligned}
\end{equation}
A higher value of $\beta$ indicates more polarization. Therefore, the polarization-based ranked topic list $f(D^L, D^R, T)$ is computed based on the corresponding polarization scores $( \beta(D^L, D^R, t_i) )_{i=1}^{K}$.

\section{Experiments and Results}

\subsection{Dataset}
We use the AYLIEN COVID-19 dataset\footnote{https://aylien.com/blog/free-coronavirus-news-dataset} consisting of \textasciitilde$1.5$M news articles related to the pandemic spanning from Nov 2019 to July 2020 that are from \textasciitilde440 global sources. To discover the polarization between politically divided news media, we select six well-known US publishers evenly split between partisan leanings: CNN, Huffington Post (Huff), New York Times (NYT) as liberal sources vs.  Fox, Breitbart (Breit) and New York Post (NYP) as conservative sources.
After filtering the publishers and remove duplicate articles, 66,368 articles are left spanning from Jan 2020 to July 2020. The statistics of news articles are shown in Appendix A.

\subsection{Experimental Setup}
\textbf{Data Preprocessing.} We build a global vocabulary containing unigrams and bigrams from the six news sources. We perform lemmatization via SPACY and remove stopwords via NLTK, where we enrich the stopwords set with ``cnn'', ``fox'', ``huffington'', and ``breitbart'' since they can bias the language model's predictions during finetuning. We desire the partisanship classification of the language model to be based on the understanding of partisanship, rather than the occurrences of news source names in the news text.

\textbf{LDA Topic Modeling.} 
We train the topic model using articles from all six sources to create a global topic set. 
The number of topics $K$ is selected from a grid search in $[10, 50]$ and the model with $K=39$ produces the best coherence value \cite{roder2015exploring}. From the 39 topics we remove 9 of them regarding advertisements, sport events, gossip news and recipes, and 30 topics are left; the removed topics are more factual and contain less ideologies from the news media, which is less worth studying.
Different from \cite{demszky2019analyzing} that assigns only one topic with the highest probability to a document, we allow a document to be assigned multiple topics with different probabilities. We represent each topic with its top-10 keywords because given a topic $t_i$ we empirically find that  $\sum_{j=1}^{10} p_{ij}>0.95$; and we keep the top-10 most relevant documents to represent a topic because 
on some topics, the documents beyond the top-10 list are obviously irrelevant and will bias the polarization study regarding the topic.
In Table \ref{tab:keywords} we show the top-$10$ keywords of topics that are discussed in this paper. For a complete list of topics please refer to Appendix B.

\begin{table*}[ht]
\centering
\begin{small}
\begin{tabular}{cl}
\hline
Idx & Top-10 keywords (and two defined stances)                                                                                                                                                                                       \\ \hline
\bf 1   & \begin{tabular}[c]{@{}l@{}}\textit{keywords}: police, officer, man, black, protest, people, arrest, kill, protester, matter\\ \textit{stances}: \textbf{protests are for social justice} vs. \textbf{protests are riots}\end{tabular}                                          \\ \cline{2-2} 
\bf 2   & \begin{tabular}[c]{@{}l@{}}\textit{keywords}: coronavirus, pandemic, federal, supply, government, make, effort, ventilator, response, agency\\ \textit{stances}: \textbf{healthcare supplies are in good condition} vs. \textbf{shortage of supplies}\end{tabular} \\ \cline{2-2} 
6   & \textit{keywords}: case, report, number, death, health, coronavirus, confirm, official, accord, covid                                                                                                                                                 \\ \cline{2-2} 
\bf 8   & \begin{tabular}[c]{@{}l@{}}\textit{keywords}: state, order, reopen, county, california, governor, business, open, jersey, guideline\\ \textit{stances}: \textbf{pro-lockdown} vs. \textbf{anti-lockdown}\end{tabular}                         \\ \cline{2-2} 
\bf 9   & \begin{tabular}[c]{@{}l@{}}\textit{keywords}: post, twitter, video, facebook, tweet, social\_media, share, write, call, make\\ \textit{stances}: 
\textbf{fact-checking is helpful} vs. \textbf{fact-checking is misleading}
\end{tabular}              \\ \cline{2-2} 
\bf 10  & \begin{tabular}[c]{@{}l@{}}\textit{keywords}: trump, president, white\_house, donald, administration, fauci, coronavirus, vice, briefing, task\_force\\ \textit{stances}: \textbf{critical of white house covid briefings} vs. \textbf{defending them}\end{tabular}              \\ \cline{2-2} 
\bf 11  & \begin{tabular}[c]{@{}l@{}}\textit{keywords}: covid, dr, coronavirus, health, disease, drug, expert, risk, treatment, director\\ \textit{stances}: \textbf{drugs promoted by Trump are risky} vs. \textbf{they are helpful}\end{tabular}                                  \\ \cline{2-2} 
\bf 12  & \begin{tabular}[c]{@{}l@{}}\textit{keywords}: mr, biden, campaign, election, party, democratic, voter, joe\_biden, republican, primary\\ \textit{stances}: \textbf{endorsing Biden in Democratic primaries} vs. \textbf{endorsing Sanders}\end{tabular}                        \\ \cline{2-2} 
\bf 27  & \begin{tabular}[c]{@{}l@{}}\textit{keywords}: year, company, market, stock, price, drop, month, business, global, sale\\ \textit{stances}: \textbf{oil/stock prices are falling} vs. \textbf{the prices are going up}\end{tabular}                                                           \\ \cline{2-2} 
28  & \textit{keywords}: state, coronavirus, cuomo, florida, texas, york, governor, tuesday, week, monday                                                                                                                                                 \\ \cline{2-2} 
29  & \textit{keywords}: house, coronavirus, republican, member, bill, senate, democrat, wednesday, washington, thursday                                                                                                                                  \\ \cline{2-2} 
\bf 30  & \begin{tabular}[c]{@{}l@{}}\textit{keywords}: country, lockdown, government, coronavirus, measure, people, italy, restriction, travel, border\\ \textit{stances}: \textbf{closing borders in Europe} vs. \textbf{opening borders} \end{tabular}    \\ \cline{2-2} 
31  & \textit{keywords}: claim, court, judge, law, federal, district, rule, chicago, legal, decision                                                                                                                                              \\ \cline{2-2} 
\bf 33  & \begin{tabular}[c]{@{}l@{}}\textit{keywords}: hospital, care, health, patient, medical, covid, center, facility, home, doctor\\ \textit{stances}: \textbf{overwhelmed hospitals} vs. \textbf{hospitals not overwhelmed}\end{tabular}                                      \\ \hline
\end{tabular}
\end{small}
    \caption{The keywords of topics discussed in the paper and two political stances of 10 labeled topics. The indices of labeled topics are highlighted in bold.}
    \label{tab:keywords}
\end{table*}

\textbf{Learning Partisanship.}
We finetune the pretrained bert-base-uncased model from huggingface Transformers \cite{wolf2020transformers} to classify the news articles according to their political leanings, or partisanship. To smooth over the differences in style and writing between the sources and render the model primarily sensitive to political divisions, we aggregate CNN, Huff, and NYP to create a holistic Liberal corpus, and similarly aggregate Fox, Breit and NYP to create a holistic Conservative corpus and optimize the model to classify whether an article is from Liberal or Conservative.
In fact, finetuning a BERT model to recognize differences only between CNN vs. Fox is likely to make it end up capturing the writing style differences and ignoring political differences, since the former is an easier task. For more details about the training process please refer to Appendix C.

\subsection{Annotating Topic Polarization}
\label{sec:annotation}
As ground truth for the evaluation of PaCTE, we annotate the topic polarization scores on a subset of the 30 modeled topics. 

We asked three annotators to select 10 topics and define two polarized political stances on each selected topic, and they reached an agreement on $T^{\text{labeled}}=\{t_1,t_2,t_8,t_9,t_{10},t_{11},t_{12},t_{27},t_{30},t_{33}\}$, as shown in Table \ref{tab:keywords}. Then on each topic in $T^{\text{labeled}}$, we selected 60 relevant documents (10 from each of the six sources), and asked three annotators to decide which stance they belong to (label it as $0/1$). If the document does not have a clear stance, it was labeled as $-1$. On each document, the majority label from the annotations was used as the final annotation. Please refer to Appendix D for more details about the annotation process.

Denoting the number of negative labels ($0$) and positive labels ($1$) in corpus $D$ on topic $t$ as $\text{N}_D^{t}(0)$ and $\text{N}_D^{t}(1)$ respectively,
the leaning of the corpus on the topic is quantified as 
\begin{equation}
    \text{le}(D, t) = (\text{N}_D^{t}(1) - \text{N}_D^{t}(0))/|D| \in [-1,1].
\end{equation}
Intuitively, $\text{le}(D,t)$ reflects how much the corpus is aligned with the stance labeled as $1$. Notably, the documents labeled with $-1$ are not counted because they do not display a clear political standing.
Accordingly, the ground-truth 
polarization score between a liberal corpus $D^L$ and a conservative corpus $D^R$ on topic $t$ is computed as the difference 
between the leanings of the two corpora, such that
\begin{equation}
 \scalebox{0.85}{$\alpha(D^L,D^R,t) = |\text{le}(D^L, t)-\text{le}(D^R, t)|/2 \in [0,1].$} 
\end{equation}

A higher value of $\alpha$ signifies more polarization. As a result, the ground-truth polarization-based topic ranked list $l_{\text{gt}}(D^L, D^R, T^{\text{labeled}})$ between a liberal corpus $D^L$ and a conservative corpus $D^R$ is computed based on the corresponding ground-truth polarization scores $( \alpha(D^L, D^R, t)  | t \in T^{\text{labeled}} )$. 


\subsection{Baselines}
We compare PaCTE to the following three baselines.

\textbf{Leave-out estimator (LOE).} For a pair of news corpora $D^L$ and $D^R$ and a given topic $t$, we take the top-10 most relevant documents from each corpus and feed the token frequency vectors of the documents into the leave-out estimator \cite{demszky2019analyzing}, from which we use estimated partisanship as the polarization score ($\in [0, 1]$) of topic $t$ between $D^L$ and $D^R$, following the idea of measuring within-topic polarization in their paper. Note that different from their method that extracts topic using embedding-based topic assignment, we use the same LDA topic model in PaCTE to extract topics, so as to ensure fair comparison between PaCTE and LOE.

\textbf{PaCTE$\neg$FT.} A variant of PaCTE without finetuning the language model. We compare to it to show the effect of finetuning the language model.

\textbf{PaCTE-PLS.} A variant of PaCTE where the language model is finetuned on news articles with partisanship labels shuffled and thus is confused about the partisanship. We compare to it in order to show the effect of rendering the language model partisanship-aware.


\subsection{Quantitative Evaluation with Labeled Topics}
\label{sec:evaluation}
To quantitatively evaluate the effectiveness of PaCTE and the baselines in capturing topic polarization, we use the 10 manually labeled topics to create a ground truth ranking of polarized topics and score models on their ability to retrieve the most polarized topics on this ranked list. 

\textbf{Evaluation protocol.} Given a liberal news corpus $D^L$, a conservative news corpus $D^R$, and a list of 10 topics ranked by ground-truth polarization scores, $l_{\text{gt}}(D^L, D^R, T^{\text{labeled}})$, as described in Section \ref{sec:annotation}, we define the top-3 topics in the list as the target polarized topics
that deserve more attention and that should be addressed when trying to prevent polarization from escalating. The target polarized topics between different pairs of news sources are shown in Table \ref{tab:pola-topics}.
Then, given a ranked list of topics $f_{\text{pred}}(D^L, D^R, T^{\text{labeled}})$ predicted by a model, we evaluate how 
effectively the 3 target polarized topics are retrieved in this model predicted list
using recall@3. 
In other words, we check how much the overlap is between the top-3 topics in the ground-truth ranking and the top-3 topics in the predicted ranking, of the 10 labeled topics. We call this task \emph{polarized topics retrieval}. 

\begin{table}[ht]
\centering
\small
\begin{tabular}{c|ccc}
\hline
              & \textbf{Fox} & \textbf{Breit} & \textbf{NYP} \\ \hline
\textbf{CNN}  & 1,9,10        & 9,1,11          & 9,10,2        \\
\textbf{Huff} & 10,1,8     & 1,11,9       & 10,12,30     \\
\textbf{NYT}  & 10,33,1     & 11,1,33       & 11,9,10    \\ \hline
\end{tabular}
\caption{The target polarized topics between different pairs of news sources from human annotations.}
\label{tab:pola-topics}
\end{table}

\textbf{Analysis of results.} The results of polarized topics retrieval using different methods in nine news corpus pairs are shown in Table \ref{tab:res-pola-retrieve}. The average recall@3 over the nine news source pairs is $0.26$, $0.04$, $0.26$, and $0.52$ on LOE, PaCTE$\neg$FT, PaCTE-PLS, and PaCTE respectively, where PaCTE outperforms all other baselines. 

\begin{table*}[ht]
\centering
\small
\addtolength{\tabcolsep}{-3pt}
\begin{tabular}{ccccc|cccc|cccc}
\hline
     & \multicolumn{4}{c|}{Fox}            & \multicolumn{4}{c|}{Breit}          & \multicolumn{4}{c}{NYP}            \\
     
     & {\scriptsize LOE} & {\scriptsize PaCTE$\neg$FT} & {\scriptsize PaCTE-PLS} & {\scriptsize PaCTE} & {\scriptsize LOE} & {\scriptsize PaCTE$\neg$FT} & {\scriptsize PaCTE-PLS} & {\scriptsize PaCTE} & {\scriptsize LOE} & {\scriptsize PaCTE$\neg$FT} & {\scriptsize PaCTE-PLS} & {\scriptsize PaCTE} \\ \hline
CNN  & 1/3   & 0        & 0         & 1/3     & 1/3   & 0        & 1/3        & 1/3     & 0   & 1/3        & 1/3         & 2/3     \\
Huff &  1/3   &     1/3     &     1/3      &  2/3     &  2/3   &     0     &     1/3      &    1/3   &  0   &    0      &     1/3      &    2/3   \\
NYT  &   1/3  &      0    &      1/3     &   1    &   1/3  &     0     &     1/3      &    1/3   &  0   &     1/3     &    0       &   1/3   \\ \hline
\end{tabular}
\addtolength{\tabcolsep}{3pt}
\caption{Recall@3 on polarized topics retrieval in nine partisan news source pairs using different methods, where we use the polarization-based topic ranked list from a model predictions $f_{\text{pred}}(D^L, D^R, T^{\text{labeled}})$ to retrieve the top-3 topics from the ground-truth ranked list $l_{\text{gt}}(D^L, D^R, T^{\text{labeled}})$. The row represents the liberal source and the column represents the conservative source in the news source pair.}
\label{tab:res-pola-retrieve}
\end{table*}

Comparing the results of LOE and PaCTE, we see that in most pairs PaCTE outperforms or ties with LOE. We argue that the inferior performance of LOE stems from its inability to capture document semantics due to the use of word frequency vectors. 
For example, in Huff vs. NYP, topic 12 is one of the target polarized topics, where documents from both stances spend the bulk of the content on the fact about the primaries and then use a few words to explicitly or implicitly endorse Biden or Sanders. Based on the use of words it is difficult to differentiate documents from the two stances, leading to the failure of LOE. In contrast, PaCTE is able to capture the contextual semantics in addition to the statistics of word usages. Therefore, even when word usages are statistically similar, PaCTE manages to discern the semantic difference and capture polarization.
However, in Huff vs. Breit, compared to LOE, PaCTE fails to retrieve topic 1 regarding ``black lives matter'', which is in the target polarized topics. On topic 1 Huff stresses ``justice'' where the news articles suggest ``police knelt on a black man'', while Breit stresses ``riot'' where the articles suggest ``the protesters loot stores and attack police''. As a result, the word usages of the articles from two stances are significantly different, which is trivial for LOE to capture, and thus LOE ranks topic 1 in a high place in the output list. Despite the difference in word usages, articles from both sources mention ``protests'' and ``violence'' a lot and their ``negative'' semantics is captured by PaCTE, leading to the perceived less polarization by PaCTE.

The worst-performing method is PaCTE$\neg$FT where the language model is not finetuned. On all topics and in all partisan news source pairs, the polarization scores given by PaCTE$\neg$FT are below 0.1 (the full range is $[0,1]$) which indicates significant alignment. However, this is contradictory to the well-known polarization in news media. Such a phenomenon demonstrates the necessity of fitting a language model on the target corpus before apply cosine similarity between learned embeddings as a measure of word and topic similarities.

In PaCTE-PLS the language model is finetuned on shuffled partisan labels that do not represent real partisanship. Compared to PaCTE$\neg$FT where the model is not finetuned at all, the performance of PaCTE-PLS improves significantly, achieving the performance on a par with LOE. However, neither PaCTE$\neg$FT nor LOE makes use of information about news partisanship, and compared to PaCTE where partisanship information is leveraged, they are still outperformed. 

\textbf{Insights into partisanship learning.} We observe that PaCTE, which is finetuned on partisanship labels, outperforms PaCTE$\neg$FT and PaCTE-PLS. We hypothesize that during the finetuning process of PaCTE, whereas the direct objective is to separate documents based on partisanship labels, the model implicitly learns the two political stances on each topic in an automatic manner; just like in human annotating, the annotators were given two groups of documents from two partisan lines, and the annotators were able to discover the two political stances after reading the documents.
Therefore, after finetuning, while the model differentiates document embeddings based on partisan divisions, it separates DC topic embeddings according to the implicitly and automatically learned political stances, bearing resemblance to human annotators' defining two political stances for topics.  As a result, we can use the partisanship-aware model to capture topic polarization arising from the partisan divisions.

\subsection{Qualitative Analysis with All Topics}
In Section \ref{sec:evaluation} we quantitatively demonstrate the effectiveness of PaCTE in retrieving polarized topics when evaluating with the 10 labeled topics. We believe that such success generalizes to the case where the input to the model is the complete topic list $T$ containing 30 topics.
In this section, we conduct a case study and retrieve the top-3 most polarized topics from $T$ in CNN vs. Fox, Huff vs. Breit and NYT vs. NYP, by PaCTE. Since we do not have the ground-truth target polarized topics from $T$, for the retrieved topics, we conduct manual inspections on relevant documents and give explanations about the polarization. For the topics in $T^\text{labeled}$, the polarization is formed due to the two political stances. Therefore in this section we only focus on the retrieved topics not in $T^\text{labeled}$.


\textbf{CNN vs. Fox.} The retrieved top-3 topics are topic 28, 6, 10, where topic 10 is in $T^{\text{labeled}}$.
The first retrieved topic is topic 28, where CNN suggests the surge of new COVID cases every day but Fox suggests that the state should reopen. On topic 6 CNN reports the serious situation of coronavirus in the US, including the high number of cases and collapse of quarantine hotels, but Fox focuses more on worldwide coronavirus situation and suggests the high number of cases in Michigan is misleading. 

\textbf{Huff vs. Breit.} The retrieved top-3 topics are topic 29, 9, 31, where topic 9 is in $T^{\text{labeled}}$.
On topic 29, Huff advocates Pelosi's coronavirus bills while Breit criticizes them.
On topic 31, the articles talk about different court cases; however, no clear polarization is discerned between the pair of news sources by manual inspections. We regard it as a failure case of PaCTE. Although the relevant articles are regarding the same topic, they have different subjects or events, and thus misleading PaCTE to perceive polarization between them. 

\textbf{NYT vs. NYP}. The retrieved top-3 topics are topic 28, 12, 10, where topic 12 and 10 are in $T^{\text{labeled}}$.
On topic 28, just as in CNN vs. Fox, NYT takes the pandemic more seriously and NYP suggests reopening. 

As a result, despite a minor error, PaCTE manages to retrieve polarized topics from $T$ on the three pairs of news sources. Although we are not able to verify if the retrieved topics are indeed the ground-truth top-3 most polarized topics, we argue that if given the ground-truth ranking on $T$, PaCTE will retain its satisfactory quantitative performance in retrieving polarized topics.

\subsection{Ablation Study: Document Embedding vs. DC Topic Embedding}
In Section \ref{sec:emb-gene} we propose to use the DC topic embedding to represent the ideology of a document on a topic, instead of using the holistic document embedding. In this section we study the difference between them.
We denote the variant of PaCTE that uses document embeddings ([CLS] token embeddings) as PaCTE-DE. First, we show the results of polarized topics retrieval using PaCTE-DE and PaCTE in three partisan news source pairs in Table \ref{tab:abl-doc-emb}.

\begin{table}[ht]
\centering
\small
\addtolength{\tabcolsep}{-3pt}
\begin{tabular}{cccc}
\hline
Method   & CNN vs. Fox & Huff vs. Breit & NYT vs. NYP \\ \hline
PaCTE-DE & 0           & 0            & 0           \\
PaCTE    & 1/3         & 1/3            & 1/3         \\ \hline
\end{tabular}
\addtolength{\tabcolsep}{3pt}
\caption{Recall@3 on polarized topics retrieval using PaCTE-DE and PaCTE in three partisan news source pairs.}
\label{tab:abl-doc-emb}
\end{table}

We observe that PaCTE-DE fails to retrieve any polarized topics in all three pairs of news sources, significantly outperformed by PaCTE.
We provide more explanations on the advantages of DC topic embeddings over document embeddings from another perspective, in addition to the capability of DC topic embedding to focus more on the topic-specific semantics in a document.  We observe that the polarization scores given by PaCTE-DE in three source pairs on all topics are above 0.98 (the range is [0,1]), suggesting that all topics are highly polarized. Therefore, as the polarization scores cluster within the interval of [0.98,1], the gaps between different scores are barely discernible, in which case the output ranked list is more susceptible to random noise during the language model finetuning and is thus more unstable and erratic. However, the output polarization scores from PaCTE are more evenly distributed in [0,1], and thus are more robust to perturbations during partisanship learning; a small perturbation on a polarization score does not affect the output ranking. As a result, PaCTE enjoys a better chance to outperform PaCTE-DE.

\begin{figure}[ht]
     \centering
     \begin{subfigure}[b]{0.237\textwidth}
         \centering
         \includegraphics[width=\textwidth]{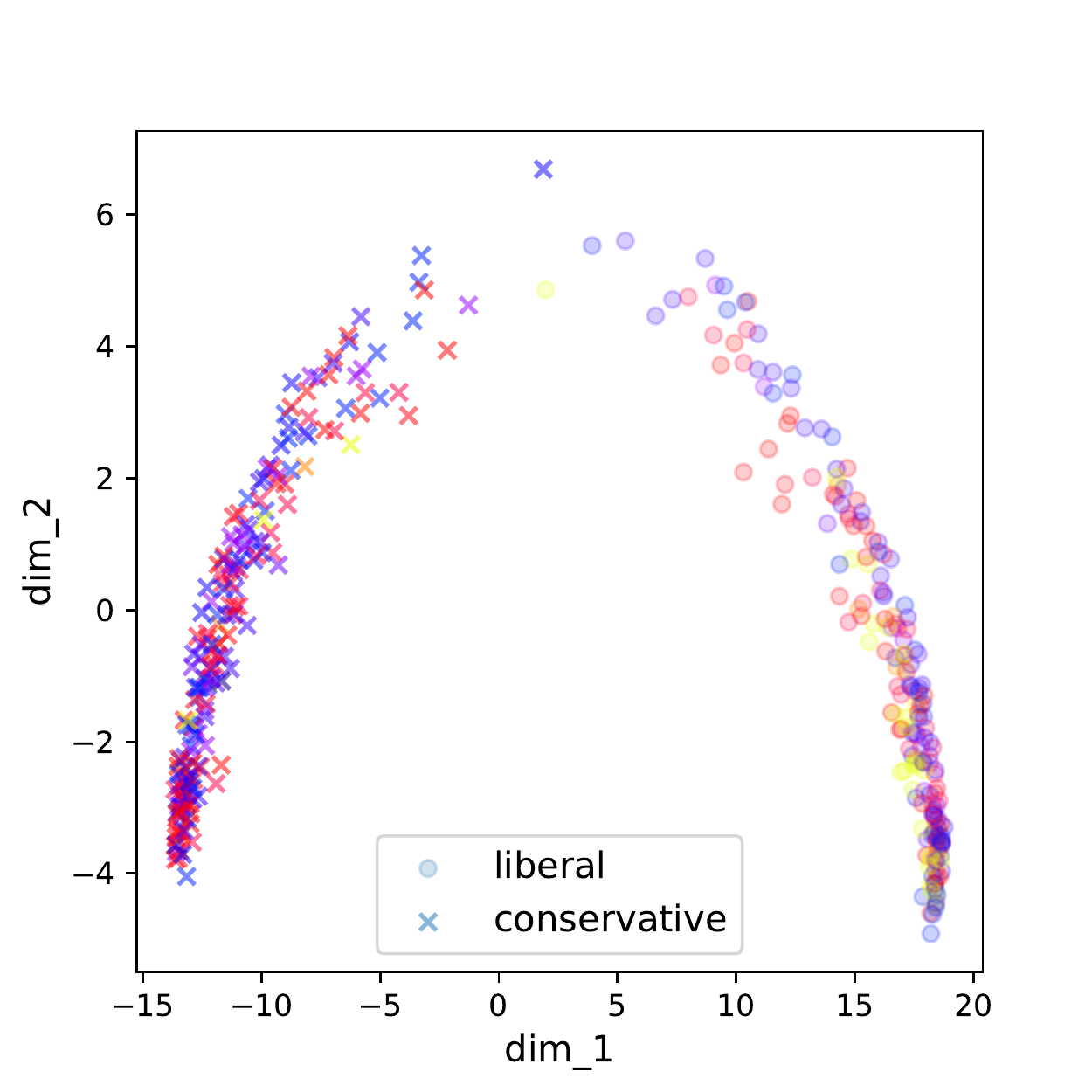}
     \end{subfigure}
     \hfill
     \begin{subfigure}[b]{0.237\textwidth}
         \centering
         \includegraphics[width=\textwidth]{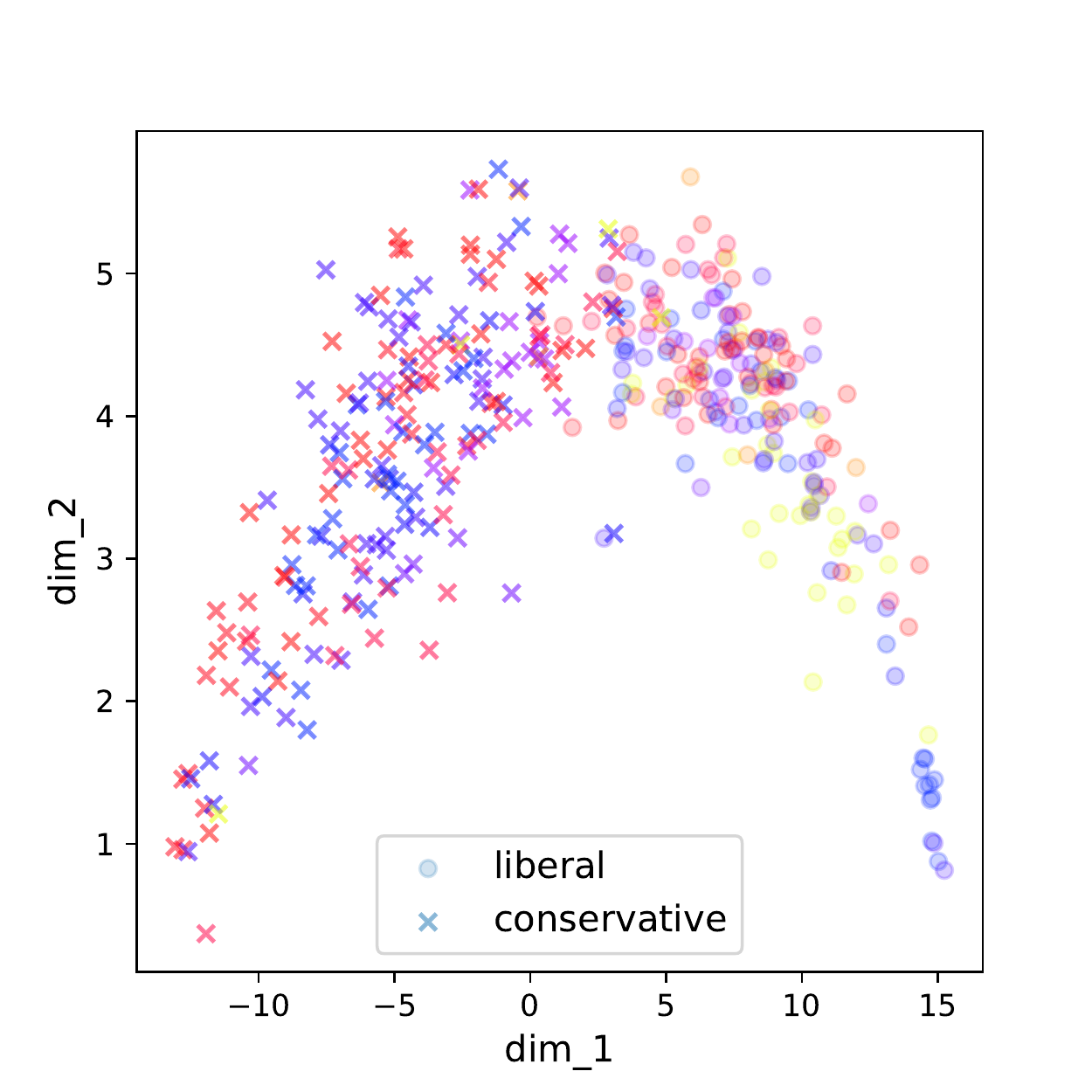}
     \end{subfigure}
        \caption{ Document embeddings (Left) and DC topic embeddings (Right) on 10 labeled topics in Liberal vs. Conservative. Different colors represent documents categorized to different topics. The original 768-d embeddings are projected into the 2-d space via PCA.}
        \label{fig:emb}
\vspace{-0.5cm}
\end{figure}

As a matter of fact, the large polarization scores from PaCTE-DE on all topics are expected, because the language model is finetuned to directly separate the document embeddings according to partisan line divisions, resulting in low cosine similarities between document embeddings on every topic, as shown in Figure \ref{fig:emb}(Left). However, despite the prominent separation of document embeddings, the corresponding DC topic embeddings that are used in PaCTE display more alignment, as shown in Figure \ref{fig:emb}(Right), where we see on some topics the DC topic embeddings are separated while on other topics the embeddings are more close. Thus, we argue that during the finetuning process, on a given topic, DC topic embeddings retain their similarity if the two partisan news articles agree on this topic, because in these articles the topic-related semantics does not contribute to the forming of the partisanship and thus maintains its position during partisanship learning, while the non-topical semantics (not captured by DC topic embeddings but captured by document embeddings) that contribute to the document partisanship keeps moving apart in the embedding space.

\section{Conclusions and Future Work}
In this paper, we propose a method to automatically discover topic-level polarization between partisan news sources by contextualized topic embeddings.
For evaluation, we create annotations on topic polarization scores in different partisan news source pairs on a variety of topics.
Compared to the leave-out estimator \cite{demszky2019analyzing} that is purely based on statistical features, our method can more precisely and meaningfully capture topical polarization as indicated by the performance on polarized topics retrieval. We hope that more NLP and researchers and practitioners can contribute to this research area that is promising but receiving insufficient attention.

Because detecting polarized topics between partisan news sources is a less established task in the research community, we articulate the data annotation and the model evaluation in great detail and make the method seemingly "complicate". However, we believe that for public media watchdogs and social media platforms to flag the highly polarized topics, our method is simple to implement, because each of the five steps described in Section \ref{sec:method} is based on robust methods in NLP. 

For future work, we plan to perform our method on more datasets, such as the tweets with noisy texts \cite{demszky2019analyzing}. In addition, we will study how to finetune the language model when when partisanship labels are not available. 

\section*{Acknowledgements}
This project was funded in part by DARPA under contract HR001121C0168. The authors are also grateful to Ves Stoyanov for a productive discussion.

\bibliography{anthology,custom}
\bibliographystyle{acl_natbib}

\clearpage
\appendix

\section{Data Preprocessing}
The statistics of the dataset is in Table \ref{tab:num-docs}. We use the summary of each news article to perform the textual analysis, because the summary contains sufficient information to understand the political stance of the article and the whole text is lengthy for the pretrained language model to handle. 
For a complete list of all documents, please check our public repository\footnote{\url{https://github.com/ZagHe568/pacte-polarized-topics-detection}}.

\begin{table*}[ht]
\centering
\begin{tabular}{l|llllllll}
              & \textbf{Jan} & \textbf{Feb}  & \textbf{Mar}   & \textbf{Apr}   & \textbf{May}   & \textbf{Jun}  & \textbf{Jul} & \textbf{\#sum} \\ \hline
\textbf{CNN}   & 232          & 589           & 2958           & 3293           & 2444           & 1789          & 2120         & \textit{13425} \\
\textbf{Fox}   & 156          & 504           & 3938           & 5148           & 3616           & 2367          & 2585         & \textit{18314} \\
\textbf{Huff}  & 17           & 74            & 1237           & 1450           & 1185           & 701           & 828          & \textit{5492}  \\
\textbf{Breit} & 93           & 240           & 1918           & 2164           & 1353           & 777           & 924          & \textit{7469}  \\
\textbf{NYT}   & 94           & 369           & 2117           & 2177           & 1730           & 849           & 1097         & \textit{8423}  \\
\textbf{NYP}   & 144          & 405           & 3063           & 3470           & 2278           & 1733          & 2152         & \textit{13245} \\
\textbf{\#sum} & \textit{736} & \textit{2171} & \textit{15231} & \textit{17702} & \textit{12606} & \textit{8216} & \textit{736} & \textit{66368}
\end{tabular}
\caption{The number of news articles from all sources in all months.}
\label{tab:num-docs}
\end{table*}

\section{LDA Topic Modeling}
We use MALLET\footnote{https://www.machinelearningplus.com/nlp/topic-modeling-gensim-python/} topic modeling. The top-10 keywords of all 39 topics are shown in Table \ref{tab:all-topics}. Among them topic 0, 3, 4, 14, 16, 26, 35, 36, 37 are not used in further analysis because after reading relevant articles we find that they are more about advertisements, sport events, gossip news and recipes and etc., which are more factual and convey limited media ideologies. 30 topics are left after removing the 9 topics. 
Table \ref{tab:all-topics} lists the top-10 keywords of the 30 topics.

\begin{table*}[ht]
\begin{small}
\begin{tabular}{c|l}
\hline
Idx & Top-10 Topic Keywords                                                                                         \\ \hline
1           & police, officer, man, black, protest, people, arrest, kill, protester, matter                           \\
2           & coronavirus, pandemic, federal, supply, government, make, effort, ventilator, response, agency          \\
5           & coronavirus, virus, test, covid, people, tested\_positive, testing, positive, symptom, spread           \\
6           & case, report, number, death, health, coronavirus, confirm, official, accord, covid                      \\
7           & event, plan, june, announce, due, july, hold, cancel, pandemic, date                                    \\
8           & state, order, reopen, county, california, governor, business, open, jersey, guideline                   \\
9           & post, twitter, video, facebook, tweet, social\_media, share, write, call, make                          \\
10          & trump, president, white\_house, donald, administration, fauci, coronavirus, vice, briefing, task\_force \\
11          & covid, dr, coronavirus, health, disease, drug, expert, risk, treatment, director                        \\
12          & mr, biden, campaign, election, party, democratic, voter, joe\_biden, republican, primary                \\
13          & school, child, student, university, parent, high, kid, year, family, class                              \\
15          & american, pandemic, crisis, america, nation, make, policy, job, people, economy                         \\
17          & time, world, space, launch, turn, center, long, life, leave, moment                                     \\
18          & coronavirus, report, outbreak, accord, ship, official, quarantine, military, force, iran                \\
19          & city, york, de\_blasio, mayor, resident, area, yorker, coronavirus, people, tuesday                     \\
20          & mask, people, wear, face, service, social\_distance, church, sunday, coronavirus, stay                  \\
21          & people, time, thing, good, work, make, lot, add, give, feel                                             \\
22          & department, official, national, security, fire, investigation, report, threat, call, director           \\
23          & employee, worker, company, restaurant, food, store, work, customer, business, amazon                    \\
24          & china, chinese, world, outbreak, virus, wuhan, organization, coronavirus, global, government            \\
25          & time, series, film, show, year, make, movie, live, race, set                                            \\
27          & year, company, market, stock, price, drop, month, business, global, sale                                \\
28          & state, coronavirus, cuomo, florida, texas, york, governor, tuesday, week, monday                        \\
29          & house, coronavirus, republican, member, bill, senate, democrat, wednesday, washington, thursday         \\
30          & country, lockdown, government, coronavirus, measure, people, italy, restriction, travel, border         \\
31          & claim, court, judge, law, federal, district, rule, chicago, legal, decision                             \\
32          & health, public, people, work, community, include, protect, provide, group, pandemic                     \\
33          & hospital, care, health, patient, medical, covid, center, facility, home, doctor                         \\
34          & program, pay, money, fund, economic, job, business, relief, federal, receive                            \\
38          & coronavirus, office, letter, pandemic, call, send, statement, issue, write, act                        
\end{tabular}
\end{small}
\caption{List of all the 39 topics with corresponding top keywords.}
\label{tab:all-topics}
\end{table*}

\subsection{News Article Examples}
On topic 10, we show six examples of news articles, one from each news source. For a complete list of news articles, please refer to our public repository.

\textbf{CNN}: \emph{There has been a concerted effort among aides and allies to get President Donald Trump to stop conducting the daily coronavirus briefings, multiple sources tell CNN. The briefing came a day after Trump had given a lengthy briefing to the media, at one point suggesting it might be possible to treat coronavirus by injecting people with sunlight or disinfectants. Trump asked White House coronavirus task force coordinator Dr. Deborah Birx during Thursday's briefing. A source close to the coronavirus task force said Trump was upset about the "flack" he was taking after those comments and that appears to be part of the reason why the President cut Friday's briefing short. During the earlier questioning from reporters on Friday, Trump said he was being "sarcastic" with his suggestion that people inject themselves with disinfectant, even though he was clearly being serious during Thursday's briefing.}

\textbf{Fox}: \emph{White House press secretary Kayleigh McEnany, during her first official briefing, promised that she ‘will never lie’ to the press in her new role. White House press secretary Kayleigh McEnany, during her first official briefing, promised that she "will never lie" to the press in her new role. McEnany took the podium for the first time Friday, after being tapped as White House press secretary from her post as national spokeswoman for President Trump's re-election campaign earlier this month. TRUMP NAMES KAYLEIGH MCENANY AS NEW WHITE HOUSE PRESS SECRETARY "I will never lie to you," McEnany told reporters. McEnany seemed to signal that the White House would scale back on their daily coronavirus task force briefings, which were regularly led by the president himself, and Vice President Pence, with appearances from Dr. Deborah Birx and Dr. Anthony Fauci to provide public health information.}

\textbf{Huffington Post}: \emph{President Donald Trump on Sunday tore into former President Barack Obama, calling him “an incompetent president” after Obama appeared to criticize his response to the coronavirus crisis during two commencement speeches a day earlier. Asked about Obama’s remarks, Trump told reporters on the White House lawn that he “didn’t hear it” before proceeding to bash his predecessor as “grossly incompetent.” President Trump: "[President Obama] was an incompetent president. But earlier this month, Obama reportedly bashed the Trump administration’s response to the pandemic as “an absolute chaotic disaster” during a phone call with some of his former White House aides. When a Washington Post reporter last week asked Trump to explain “Obamagate,” the president refused.}

\textbf{Breibart}: \emph{New York magazine Washington correspondent Olivia Nuzzi responded angrily to criticism from former White House press secretary Ari Fleischer on Monday evening, tweeting at him: “Oh shut the f*ck up.” Fleisher, who served under President George W. Bush, criticized Nuzzi after a Rose Garden press briefing on the coronavirus pandemic in which she asked President Donald Trump: “If an American president loses more Americans over the course of six weeks than died in the entirety of the Vietnam War, does he deserve to be re-elected?” One example is a “fake news” viral photograph of President Lyndon B. Johnson, which was presented by many Trump critics as if Johnson had been expressing grief over the deaths in Vietnam. President Trump is said to be reconsidering his daily press briefings because journalists use them to grandstand and to score political points, rather than to pursue information. The contrast with press briefings for governors and mayors is stark: there, journalists tend to be more deferential and to ask questions aimed at eliciting information rather than assigning political fault.}

\textbf{New York Times}: \emph{WASHINGTON — After several days spent weathering attacks from White House officials, Dr. Anthony S. Fauci hit back on Wednesday, calling recent efforts to discredit him “bizarre” and a hindrance to the government’s ability to communicate information about the coronavirus pandemic. On Wednesday, Peter Navarro, Mr. Trump’s top trade adviser, published a brazen op-ed article in USA Today describing Dr. Fauci as “wrong about everything.” All the while, White House officials — including the president and the press secretary — assert in the face of the evidence that there is no concerted effort to attack Dr. Fauci. He has not briefed Mr. Trump in weeks, preferring to work with Dr. Deborah L. Birx, who helps coordinate the administration’s coronavirus response, or to send his messages through Vice President Mike Pence. In the piece, Mr. Navarro presented what White House officials have been saying privately about Dr. Fauci, and what Mr. Trump has said publicly: They like Dr. Fauci personally, but he has made mistakes.}

\textbf{New York Post}: \emph{President Trump said Wednesday he has a “very good relationship” with White House coronavirus task force member Anthony Fauci, despite an op-ed by one of his top advisers that trashed the immunologist. Trump distanced himself from trade adviser Peter Navarro’s op-ed that said Fauci “has been wrong about everything.” “I get along very well with Dr. Fauci,” Trump told reporters in the Oval Office. On Monday, White House Press Secretary Kayleigh McEnany denied a Washington Post report that said reporters were given “opposition research” to discredit Fauci, including his past remarks early on in the pandemic that the public didn’t need to wear masks. “We were asked a very specific question by the Washington Post, and that question was President Trump noted that Dr. Fauci had made some mistakes, and we provided a direct answer to what was a direct question.”}

\subsection{Top-10 Most Relevant Documents on All Topics}
We show the topic-10 most relevant document indices on all 30 topics on each source. On some topics there are less than 10 relevant documents on some sources. Note that such topics are not in the 10 labeled topics and are only used for qualitative analysis; in other words, for quantitative analysis, we ensure that on all the 10 labeled topics, there are 10 relevant documents on each source.

\textbf{Topic 1.} 
\textbf{CNN}: 22873, 62724, 62635, 63979, 60318, 64323, 39686, 23087, 66007, 64346. 
\textbf{Fox}: 64455, 21485, 26889, 21509, 22055, 62291, 21458, 22866, 21404, 65938. 
\textbf{Huff}: 22937, 63328, 40375, 66026, 64381, 61909, 64511, 63335, 64173, 64573. 
\textbf{Breit}: 64348, 52383, 21375, 32945, 64143, 58746, 65060, 37841, 21378, 40036. 
\textbf{NYT}: 22742, 65966, 21562, 58360, 21357, 65875, 65146, 21330, 52138, 21501.
\textbf{NYP}: 21316, 62547, 64164, 21638, 64765, 7055, 21590, 60740, 21720, 40400.

\textbf{Topic 2.} 
\textbf{CNN}: 2291, 9180, 42882, 6582, 1629, 1476, 11612, 850, 42172, 2006.
\textbf{Fox}: 9182, 31226, 2682, 48284, 10207, 46558, 11722, 10312, 3984, 11122.
\textbf{Huff}: 889, 35891, 48378, 46288, 5361, 708, 2408, 1319, 1938, 8962.
\textbf{Breit}: 3817, 13126, 6136, 55780, 13045, 3958, 17769, 9004, 48430, 18855.
\textbf{NYT}: 3723, 44895, 12794, 54586, 8385, 61651, 33770, 51084, 33735, 11857.
\textbf{NYP}: 46244, 46842, 8039, 62990, 1877, 17378, 66227, 786, 39335, 809.

\textbf{Topic 5.}
\textbf{CNN}: 52276, 22960, 63870, 63867, 9240, 22872, 56452, 28994, 13558, 52074.
\textbf{Fox}: 4863, 14561, 13454, 43376, 14939, 8762, 12185, 8924, 14800, 6561.
\textbf{Huff}: 12973, 29357, 5156, 12938, 2941, 22574, 14862, 43495, 29785, 29580.
\textbf{Breit}: 13167, 64051, 8622, 12110, 14544, 62240, 51507, 15519, 18759, 47388.
\textbf{NYT}: 29543, 19969, 2964, 30581, 49542, 24286, 61430, 35059, 4465, 66042.
\textbf{NYP}: 15340, 13404, 54124, 64067, 35789, 9325, 22692, 3575, 8217, 15513.

\textbf{Topic 6.}
\textbf{CNN}: 53265, 56003, 16985, 15115, 26006, 17063, 19812, 50074, 18014, 21857.
\textbf{Fox}: 58537, 63076, 53236, 22482, 50794, 42788, 60553, 59539, 63026, 54784.
\textbf{Huff}: 54194, 37337, 64330, 35448, 10906, 6077, 48061, 48797, 20643, 25306.
\textbf{Breit}: 36565, 61077, 44474, 49577, 17068, 48689, 14186, 16924, 18078, 29693.
\textbf{NYT}: 62593, 50117, 28572, 39978, 7679, 56428, 5277, 23283, 18404, 20709.
\textbf{NYP}: 62856, 66347, 48582, 60094, 6419, 54128, 21187, 59736, 60922, 59286.

\textbf{Topic 7.}
\textbf{CNN}: 8968, 47625, 14093, 8928, 10954, 17954, 12589, 3657, 29776, 54903.
\textbf{Fox}: 14933, 64745, 50666, 8112, 48177, 63686, 8539, 16974, 10530, 62301.
\textbf{Huff}: 43539, 45395, 41053, 41201, 4411, 14668, 62437, 25455, 17322, 33304.
\textbf{Breit}: 14659, 16032, 15822, 13584, 4032, 8559, 16522, 6377, 48135, 45140.
\textbf{NYT}: 36774, 1098, 41903, 6893, 22100, 42130, 40933, 65242, 13346, 12585.
\textbf{NYP}: 50435, 61426, 4653, 48992, 41460, 41187, 49057, 17932, 59701, 59289.

\textbf{Topic 8.}
\textbf{CNN}: 52753, 35955, 53068, 52474, 8618, 36939, 58294, 53262, 54031, 39671.
\textbf{Fox}: 52787, 36759, 28989, 41859, 38546, 35214, 10486, 51718, 32366, 38501.
\textbf{Huff}: 52782, 13488, 33939, 26489, 21984, 35994, 58843, 24991, 22420, 54017.
\textbf{Breit}: 36756, 25982, 26581, 56836, 31828, 31625, 24778, 54023, 29688, 22190.
\textbf{NYT}: 26442, 10769, 28108, 54705, 12483, 8496, 33668, 29239, 34964, 25184.
\textbf{NYP}: 54049, 32336, 60474, 10483, 25023, 10508, 58551, 62855, 22591, 6424.

\textbf{Topic 9.}
\textbf{CNN}: 24337, 29633, 49300, 18255, 21561, 63481, 65460, 12850, 22029, 5379.
\textbf{Fox}: 58858, 22378, 24954, 10605, 49169, 28970, 42046, 22859, 17031, 22101.
\textbf{Huff}: 24163, 22956, 28908, 20469, 62999, 28039, 22394, 64592, 22044, 49612.
\textbf{Breit}: 58881, 63178, 22588, 51608, 11734, 23200, 63795, 65340, 36749, 49565.
\textbf{NYT}: 25897, 22208, 65768, 63558, 57615, 61546, 51728, 56320, 16812, 26196.
\textbf{NYP}: 29025, 53038, 11502, 17723, 49458, 37686, 53745, 24250, 54854, 24424.

\textbf{Topic 10.}
\textbf{CNN}: 36977, 9900, 60315, 47434, 52901, 61685, 15586, 33393, 16909, 28711.
\textbf{Fox}: 33421, 52296, 33431, 33349, 5564, 15986, 46620, 9827, 24355, 53249.
\textbf{Huff}: 27192, 23019, 37029, 46903, 8395, 30295, 38470, 8936, 32619, 33415.
\textbf{Breit}: 35799, 54043, 54046, 43323, 47227, 47486, 61645, 18345, 13315, 31312.
\textbf{NYT}: 53227, 26542, 43823, 33269, 52807, 50983, 10166, 33072, 52406, 14937.
\textbf{NYP}: 43755, 53340, 43996, 51619, 58930, 16594, 32531, 48513, 60137, 37131.

\textbf{Topic 11.}
\textbf{CNN}: 43064, 24419, 43853, 37234, 24293, 62395, 60856, 37716, 65640, 11503.
\textbf{Fox}: 6852, 44149, 42764, 26980, 37265, 34316, 65050, 28248, 42912, 22639.
\textbf{Huff}: 10973, 24477, 33168, 30443, 65264, 62143, 32735, 23337, 48333, 38400.
\textbf{Breit}: 4390, 62298, 46481, 10710, 21938, 44850, 11701, 378, 5428, 8059.
\textbf{NYT}: 44316, 24463, 45811, 37043, 24655, 39125, 36073, 338, 26193, 33556.
\textbf{NYP}: 46561, 62407, 65900, 7878, 10542, 52025, 61086, 34882, 43578, 19746.

\textbf{Topic 12.}
\textbf{CNN}: 16723, 17545, 13563, 12103, 14653, 42843, 25659, 35193, 15965, 14021.
\textbf{Fox}: 34526, 25672, 37152, 16593, 12336, 11898, 27791, 15679, 64568, 43431.
\textbf{Huff}: 12071, 36383, 63407, 12041, 39262, 15806, 43507, 53503, 35194, 16040.
\textbf{Breit}: 21477, 45924, 15829, 15787, 60055, 46496, 14115, 34548, 57445, 15883.
\textbf{NYT}: 18071, 18061, 18087, 18091, 18574, 18109, 18044, 2961, 18097, 44621.
\textbf{NYP}: 13572, 15798, 63085, 6355, 46392, 44686, 16507, 32856, 13753, 59949.

\textbf{Topic 13.}
\textbf{CNN}: 52042, 53647, 56269, 48533, 50908, 32977, 52157, 325, 54723, 19008.
\textbf{Fox}: 55527, 55611, 52290, 50854, 55905, 55553, 9799, 50424, 52626, 55145.
\textbf{Huff}: 57978, 10357, 12659, 57989, 28854, 15152, 48231, 39256, 41416, 50280.
\textbf{Breit}: 35932, 53281, 51224, 52932, 52933, 6724, 61389, 54136, 1500, 4253.
\textbf{NYT}: 51392, 56173, 37773, 17033, 53487, 14635, 37881, 14693, 28559, 29795.
\textbf{NYP}: 12922, 57494, 62649, 33391, 33069, 48882, 35976, 33935, 66118, 49222.

\textbf{Topic 15.}
\textbf{CNN}: 35585, 58100, 63056, 60887, 66162, 23165, 33687.
\textbf{Fox}: 64255, 57723, 61319, 47541, 57594, 60724, 46614, 65841, 32082, 51435.
\textbf{Huff}: 50283, 50286, 29216, 64189, 37388, 51475, 43624, 23605, 64125, 38169.
\textbf{Breit}: 1602, 47078, 51743, 41817, 38117, 8476, 32283, 47210, 21692, 14994.
\textbf{NYT}: 26295, 26104, 35795, 9649, 169, 53045, 26517, 49375, 10809, 64661.
\textbf{NYP}: 41409, 23782.

\textbf{Topic 17.}
\textbf{CNN}: 21394, 46193, 21796, 50681, 2467, 52824, 23060, 42194, 33416, 33996.
\textbf{Fox}: 22895, 21313, 21429, 24489, 25071, 24318, 25695, 56398, 37404, 21419.
\textbf{Huff}: 62406, 22703, 48727, 46191, 51732, 8868, 56959, 44157, 13008, 36388.
\textbf{Breit}: 22979, 48822, 21421, 60683, 39031, 24114, 43545.
\textbf{NYT}: 21423, 21486, 41181, 38254, 48811, 56904, 53586, 52196, 54239, 51030.
\textbf{NYP}: 27696, 41257, 33369, 12983, 41190, 35086, 33866, 4693, 42508, 60983.

\textbf{Topic 18.}
\textbf{CNN}: 39244, 42833, 687, 47269, 46101, 1227, 46316, 43181, 299, 37315.
\textbf{Fox}: 61023, 37051, 47122, 34309, 48330, 20285, 20295, 42259, 7247, 9568.
\textbf{Huff}: 3282, 44890, 37119, 45584, 47564, 154, 29728, 5455, 45893, 19422.
\textbf{Breit}: 124, 36851, 60988, 2273, 553, 979, 64492, 5952, 7323, 13383.
\textbf{NYT}: 173, 36987, 41045, 48109, 47310, 1555, 12502, 42449, 56665, 38772.
\textbf{NYP}: 1007, 48021, 47721, 47477, 3308, 1597, 105, 37866, 34656, 35694.

\textbf{Topic 19.}
\textbf{CNN}: 65216, 65919, 47703, 15243, 64710, 35077, 61871, 12372, 7614, 4141.
\textbf{Fox}: 1690, 12309, 65586, 29089, 3600, 29899, 55457, 22667, 62019, 55717.
\textbf{Huff}: 21497, 22538, 34750, 19739, 47734, 13671, 9080, 44803, 9790, 6579.
\textbf{Breit}: 26804, 6509, 30461, 11703, 13818, 35813, 34832, 36035, 34461, 40208.
\textbf{NYT}: 56435, 65514, 11949, 35123, 60194, 56984, 7009, 24659, 15179, 32155.
\textbf{NYP}: 31276, 64202, 33547, 36193, 61943, 6662, 50296, 41943, 37147, 12254.

\textbf{Topic 20.}
\textbf{CNN}: 29359, 41880, 44391, 23929, 23891, 58610, 16, 7791, 43279, 43686.
\textbf{Fox}: 41486, 4380, 25130, 44564, 7153, 4102, 62478, 45326, 48008, 44450.
\textbf{Huff}: 52707, 34296, 43615, 42366, 8483, 61865, 53883, 58217, 58209, 52080.
\textbf{Breit}: 40577, 50156, 44531, 1704, 224, 45886, 4078, 30607, 36544, 21342.
\textbf{NYT}: 43680, 43132, 3466, 46829, 16385, 44529, 44498, 13911, 18247, 44490.
\textbf{NYP}: 58586, 13713, 14492, 22920, 14037, 44445, 36107, 213, 4948, 62594.

\textbf{Topic 21.}
\textbf{CNN}: 30694, 38631, 10489, 64595, 46011, 58309, 26764, 39667, 30189, 62513.
\textbf{Fox}: 54100, 5720, 37070, 53083, 36870, 4434, 51052, 40273, 44540, 58646.
\textbf{Huff}: 53895, 5143, 64645, 53896, 60453, 50905, 63982, 27568, 26936, 34053.
\textbf{Breit}: 24115, 26582, 38790, 9481.
\textbf{NYT}: 2723, 27086, 52363, 10737, 19923, 44353, 6979, 56929, 37268, 8272.
\textbf{NYP}: 57065, 3175, 39506, 58484, 50308, 53214.

\textbf{Topic 22.}
\textbf{CNN}: 23167, 48096, 26205, 22047, 47742, 34234, 33189, 22970, 49527, 39375.
\textbf{Fox}: 31847, 29072, 30409, 30494, 30000, 30115, 39926, 33583, 29766, 59646.
\textbf{Huff}: 23126, 47403, 28642, 14644, 7126, 66321, 32087, 60313, 46938, 46902.
\textbf{Breit}: 35477, 46795, 27951, 59637, 38856, 46490, 33413, 13190, 6100, 28231.
\textbf{NYT}: 44318, 37139, 44335, 9489, 10618, 24316, 28351, 23162, 25276, 22511.
\textbf{NYP}: 29730, 27592, 60671, 7102, 47325, 29276, 26228, 14283, 60403, 50133.

\textbf{Topic 23.}
\textbf{CNN}: 5326, 44622, 44231, 11176, 6496, 52668, 14017, 43584, 12395, 46745.
\textbf{Fox}: 4017, 31428, 44209, 1964, 37142, 14336, 45345, 46060, 44181, 9230.
\textbf{Huff}: 12371, 3245, 49665, 49671, 6603, 10079, 52967, 59703, 1622, 5552.
\textbf{Breit}: 43555, 30686, 29565, 56249, 45016, 953, 9378, 51834, 1867, 37187.
\textbf{NYT}: 11277, 11450, 61190, 11747, 35565, 62501, 45602, 10102, 55676, 24533.
\textbf{NYP}: 3088, 43904, 42403, 40986, 12588, 31823, 12860, 43962, 45447, 60394.

\textbf{Topic 24.}
\textbf{CNN}: 11322, 12476, 51400, 32280, 26578, 7698, 22944, 53884, 22879, 27018.
\textbf{Fox}: 41909, 8147, 52862, 9992, 11884, 61000, 2816, 42300, 64408, 37554.
\textbf{Huff}: 20522, 21145, 7456, 20030, 21166, 6756, 62148, 51402, 66041, 9049.
\textbf{Breit}: 45020, 25401, 36033, 19866, 23527, 16583, 63436, 46336, 20068, 52811.
\textbf{NYT}: 15450, 30846, 19962, 5814, 46898, 33715, 2936, 17488, 18875, 32108.
\textbf{NYP}: 52593, 25600, 15066, 25704, 27623, 32851, 20066, 33527, 10198, 31849.

\textbf{Topic 25.}
\textbf{CNN}: 22679, 52783, 36356, 50190, 50458, 47910, 44222, 40460, 27182, 2412.
\textbf{Fox}: 63929, 56052, 27995, 63404, 64738, 13451, 2886, 28144, 11928, 45087.
\textbf{Huff}: 25128, 32938, 53468, 22477, 58245, 13502, 16615, 52208, 51719, 34708.
\textbf{Breit}: 43320, 58828, 3398, 14846, 51825, 13013, 19562, 49059, 56808, 49844.
\textbf{NYT}: 46, 62894, 28598, 9397, 56747, 49004, 52510, 22834, 18371, 28077.
\textbf{NYP}: 27689, 48044, 60205, 51648, 27053, 65173, 62536, 36022, 49470, 55655.

\textbf{Topic 27.}
\textbf{CNN}: 48318, 39788, 40268, 30754, 35012, 3553, 42657, 42793, 58409, 43689.
\textbf{Fox}: 10658, 16722, 34052, 16603, 16563, 38831, 12629, 3598, 16775, 3114.
\textbf{Huff}: 9897, 16530, 55262, 1735, 47491, 17807, 47369, 11247, 15041, 2377.
\textbf{Breit}: 7705, 16758, 15418, 13101, 19347, 10946, 59668, 17223, 18750, 16521.
\textbf{NYT}: 44301, 27331, 3481, 16605, 33742, 17125, 16868, 16467, 22856, 38888.
\textbf{NYP}: 58424, 62470, 58079, 66201, 33378, 34479, 19782, 16541, 41876, 16748.

\textbf{Topic 28.}
\textbf{CNN}: 31044, 44809, 60338, 8200, 52516, 22992, 1983, 48445, 58510, 55807.
\textbf{Fox}: 46514, 39209, 39519, 29768, 33664, 39207, 59484, 58943, 35452, 27794.
\textbf{Huff}: 41088, 2294, 1791, 2527, 16944, 35820, 58940, 36351, 29250, 33344.
\textbf{Breit}: 59388, 10467, 32176, 21937, 47074, 59991, 40333, 45801, 54859, 10944.
\textbf{NYT}: 3442, 2528, 36292, 57818, 26841, 79, 5783, 44555, 26664, 13881.
\textbf{NYP}: 37264, 60701, 63209, 23180, 10178, 61152, 45302, 45076, 21315, 37572.

\textbf{Topic 29.}
\textbf{CNN}: 5370, 5943, 16279, 41875, 23014, 2499, 6987, 6015, 8575, 5551.
\textbf{Fox}: 22625, 22575, 8567, 3697, 22756, 38698, 21966, 5688, 9058, 49027.
\textbf{Huff}: 5705, 3418, 5282, 39977, 878, 38365, 41605, 27619, 33204, 5578.
\textbf{Breit}: 2357, 34682, 5694, 27927, 8582, 26267, 32384, 15207, 7962, 51745.
\textbf{NYT}: 38387, 39511, 5632, 30272, 11624, 11802, 5849, 42372, 15295, 1519.
\textbf{NYP}: 38590, 46541, 43984, 39203, 22612, 8518, 3227, 5616, 35491, 6413.

\textbf{Topic 30.}
\textbf{CNN}: 12834, 12820, 56912, 2409, 57696, 29817, 50038, 27123, 65625, 56735.
\textbf{Fox}: 15271, 16197, 15482, 49437, 16316, 11890, 4167, 12690, 63882, 16313.
\textbf{Huff}: 65420, 14194, 58055, 35495, 52096, 46084, 18917, 43566, 64277, 53090.
\textbf{Breit}: 7890, 15098, 27515, 57927, 16468, 29475, 23506, 16687, 18931, 54634.
\textbf{NYT}: 64811, 14150, 11027, 9950, 30034, 62830, 49472, 57934, 12405, 8467.
\textbf{NYP}: 63157, 16513, 58020, 25525, 27500, 35402, 26423, 44011, 13333, 58093.

\textbf{Topic 31.}
\textbf{CNN}: 40883, 60073, 45442, 47780, 66073, 47018, 57685, 1530, 1803, 27870.
\textbf{Fox}: 38325, 5939, 39474, 54708, 44688, 34237, 54184, 1421, 47715, 35102.
\textbf{Huff}: 6195, 53859, 65214, 65114, 53135, 55111, 47030, 43457, 47202, 12078.
\textbf{Breit}: 58702, 57208, 51227, 25817, 27911, 40737, 55540, 18717, 42631, 62134.
\textbf{NYT}: 65623, 13394, 57574, 59446, 56216, 55016, 52453, 32816, 49563, 36481.
\textbf{NYP}: 47461, 55354, 53865, 57144, 55160, 57763, 51672, 42502, 58347, 65727.

\textbf{Topic 32.}
\textbf{CNN}: 5228, 18432, 17048, 60294, 31452, 17064, 1250, 43011, 31358, 31484.
\textbf{Fox}: 24593, 37494, 38517, 41439, 54372, 23690, 45754, 61263, 40552, 1334.
\textbf{Huff}: 5774, 34939, 2302, 8251, 50885, 15592, 696, 50917, 5142, 50876.
\textbf{Breit}: 44194, 38015, 7918, 43423, 9199, 27009, 31618, 6376, 38280, 9201.
\textbf{NYT}: 54223, 28551, 43858, 2159, 26747, 30253, 42923, 27523, 21090, 27586.
\textbf{NYP}: 37943, 58555, 26131, 59816, 42493, 37261, 33388, 31703, 6127, 40455.

\textbf{Topic 33.}
\textbf{CNN}: 4440, 46162, 47741, 41021, 38965, 22754, 12074, 39663, 50663, 7134.
\textbf{Fox}: 32866, 55982, 47482, 45951, 53253, 43766, 42071, 32533, 46577, 6228.
\textbf{Huff}: 64120, 57480, 48002, 44465, 31905, 6234, 1683, 49307, 16780, 6000.
\textbf{Breit}: 2210, 36748, 36102, 27893, 51132, 9938, 6838, 4996, 24471, 704.
\textbf{NYT}: 15462, 51790, 1032, 9831, 36054, 29577, 44683, 31507, 43101, 43299.
\textbf{NYP}: 4888, 498, 21705, 6785, 26875, 2439, 35833, 54726, 41755, 24068.

\textbf{Topic 34.}
\textbf{CNN}: 6266, 65009, 66144, 43941, 7262, 32916, 39176, 32957, 7827, 65108.
\textbf{Fox}: 54014, 37279, 25070, 25979, 57486, 2458, 55106, 53826, 38417, 42652.
\textbf{Huff}: 63169, 712, 33532, 42077, 30986, 35798, 56157, 4025, 51157, 54739.
\textbf{Breit}: 7502, 12430, 1267, 13007, 8824, 24116, 7093, 5218, 25354, 56127.
\textbf{NYT}: 47350, 46129, 28442, 46107, 58116, 37725, 2418, 33019, 43138, 30807.
\textbf{NYP}: 38718, 1997, 54988, 41888, 42854, 45033, 35790, 806, 8543, 63421.

\textbf{Topic 38.}
\textbf{CNN}: 26089, 9780, 11160, 10135, 38444, 10528, 15396, 21793, 11899, 30939.
\textbf{Fox}: 9904, 61954, 38592, 1912, 9742, 48158, 47330, 11298, 2145, 1109.
\textbf{Huff}: 1053, 10636, 5194, 56248, 59653, 10001, 16251, 26935.
\textbf{Breit}: 9820, 37006, 9757, 10047, 43153, 47998, 15249, 10466, 4255, 6684.
\textbf{NYT}: 28253, 10549, 21983, 10078, 38217, 14959, 9733, 60099, 9898.
\textbf{NYP}: 10117, 42971, 54551, 10258, 42732, 37803, 28002, 9832, 31722, 10433.

\section{Learning Partisanship}
The news articles are split into the training set comprising topical documents and validation set comprising non-topical documents. 
Non-topical documents have small probabilities (<0.15) categorized to all topics. We do such a split because all documents are assigned a partisanship label, but not all of them are topical. For the topical documents from which we will generate contextualized topic embeddings, we use them as the training data to finetune the language model during the training phase. As a result, train set has 30,571 documents and the validation set has 35,797 documents. The model is trained for $30$ epochs and we pick the one with the best performance on validation set for the subsequent topic embedding generation. We train the model using Adam optimizer, with learning rate 1e-5 and weight decay 5e-4. We use a batch size of 64 and train the model on 4 RTX 2080 GPUs. Each epoch takes about 10 minutes. The best validation F1 score on classifying partisanship is $91.3$. 

\section{Annotating Topic Polarization}
We recruit 3 annotators that work as academic researchers in the areas of NLP and social science. For each one of the 30 topics, the annotators are provided with the top-10 topic keywords and the summaries of top-10 most relevant documents from each news corpus (as a total of 60 documents). First, the annotators select 15 topics on which they feel it is straightforward to find two polarized political stances by reading the relevant documents. For example, on topic 12 about Democratic primaries, it is intuitive to perceive the two political stances are ``endorsing Biden'' and ``endorsing Sanders'' after reading relevant articles, and then this topic is likely to be selected. We take the overlap of the 15 selected topics from 3 annotators and obtain 10 topics: $T^{\text{labeled}}=\{t_1,t_2,t_8,t_9,t_{10},t_{11},t_{12},t_{27},t_{30},t_{33}\}$ with defined polarized political stances. In other words, the annotators reach an agreement that it is more clear on these 10 topics that there are two political stances. We find that on each of these 10 topics, the two stances defined by 3 annotators reach a complete agreement.

We do not annotate all topics because 1) it is difficult for humans to discern the two political stances on some topics, especially when such two stances do not exist at all; 2) we use the vanilla LDA topic modeling which is not the state-of-the-art, so the modeled topics will change using different topic models, in which case the annotating step should be repeated. Nevertheless, we argue that annotating 10 topics is sufficient to quantitatively evaluate the effectiveness of PaCTE.

Given a topic $t$ from $T^{\text{labeled}}$, the defined two stances, and its 60 most relevant documents (10 from each of the six news sources), for each document, we ask the annotators to label which stance it belongs to and label it as 0 or 1; if the annotator is not able to perceive a clear political stance, then the annotator will label it as -1. For each document, the majority vote of the three labels with be used as the final annotation. If no majority vote is achieved, in other words, the three annotators give three different labels to a document, then a fourth annotator will read the document again and decide the final label. For a complete list of all document labels on the 10 selected topics, please refer to our public repository.

\end{document}